\documentclass[letterpaper, 10 pt, conference]{ieeeconf}  % Comment this line out if you need a4paper

\IEEEoverridecommandlockouts                              % This command is only needed if 
\usepackage{times}
\usepackage{dsfont}
\usepackage{graphicx}
\usepackage{multicol}
\usepackage{multirow}
\usepackage{makecell}
\usepackage[bookmarks=true]{hyperref}
\hypersetup{
    colorlinks=true,
    linkcolor=blue,
    filecolor=magenta,      
    urlcolor=black, % cyan
    pdftitle={Overleaf Example},
    pdfpagemode=FullScreen,
    }
\usepackage[utf8]{inputenc} % allow utf-8 input
\usepackage[T1]{fontenc}    % use 8-bit T1 fonts
\usepackage{url}            % simple URL typesetting
\usepackage{booktabs}       % professional-quality tables
\usepackage{amsfonts}       % blackboard math symbols
\usepackage{amssymb}
\usepackage{amsmath} % Required for \argmin
\usepackage{nicefrac}       % compact symbols for 1/2, etc.
\usepackage{xcolor}         % colors
\usepackage{caption}
\usepackage{subcaption}

\usepackage{bbm}
\usepackage{bm}
\usepackage{lipsum}
\usepackage{algorithm}
\usepackage{algpseudocode}
\usepackage{framed}
\usepackage{siunitx}
\usepackage{bbm}

\let\labelindent\relax
\usepackage{enumitem}

\usepackage{mathtools}
\graphicspath{ {images/} }

\usepackage{svg}

\usepackage{kotex} % Eliminate before submission

\title{\LARGE \bf
Sling2Sim2Real: One-Shot Elastic System Identification \\
for Non-Destructive Slingshot Policy Learning
}

\author{Wonjae Kang\textsuperscript{*}, Geonwoo Kim\textsuperscript{*}, Minseok Song and Daehyung Park\textsuperscript{\textdagger}% <-this % stops a space
\thanks{All authors are with the School of Computing, Korea Advanced Institute of Science and Technology, Korea ({\tt\small \{k1j24, geonwookim, hjmngb, daehyung\}@kaist.ac.kr}).  
\textsuperscript{*}These authors contributed equally to this work. {\textsuperscript{\textdagger}}D. Park is the corresponding author. This work was partly supported by Institute of Information \& communications Technology Planning \& Evaluation (IITP) grants funded by the Korea government (MSIT) (No. RS-2022-II220311, RS-2024-00336738, and RS-2024-00457882), and the IITP(Institute of Information \& Coummunications Technology Planning \& Evaluation)-ITRC(Information Technology Research Center) grant funded by the Korea government(Ministry of Science and ICT)(IITP-2026-2024-00437102), and the National Research Council of Science \& Technology(NST) grant by the Korea government (MSIT) (No. GTL25042-000).
} % Closes \thanks
} % Closes \author

\begin{document}

\maketitle
\thispagestyle{empty}
\pagestyle{empty}

\begin{abstract}
Elastic object manipulation (EOM) involves high-dimensional, nonlinear, and elastic deformations. The diverse deformation properties of elastic objects substantially expand the relevant state space, requiring extensive exploration to learn accurate manipulation policies for tasks such as slingshot manipulation. While simulation enables large-scale and safe exploration compared to costly and potentially destructive real-world trials (e.g., repeated projectile launches), accurately calibrating elastic behavior between the real world and simulation remains challenging since elastic properties are largely indistinguishable from visual observations alone. To address these challenges, we propose Sling2Sim2Real, a one-shot Real2Sim2Real framework that identifies elastic parameters from a single non-destructive interaction and enables policy learning in simulation. The framework consists of two stages: 1) a multi-start Real2Sim system identification method that exploits parameter covariance to estimate elastic properties, and 2) simulation-based policy learning followed by zero-shot Sim2Real transfer using the calibrated simulator. We evaluate Sling2Sim2Real on a slingshot manipulation task using a Franka Emika Panda arm and elastic bands with diverse physical properties across varying target distances. Experimental results demonstrate that Sling2Sim2Real achieves accurate policy learning and robust generalization while significantly reducing the amount of required real-world interaction.

%In this work, we focus on Real2Sim2Real policy learning for novel elastic sling-based slingshot manipulation. 

%, where policies are learned efficiently and safely in simulation and transferred to the real world in zero shot. A key challenge in EOM arises from the visually indistinguishable nature of elastic properties, which complicates the real2sim transfer. 
%makes hard to set identical simulated target objects. This leads to the gap between trained policies in simulation and real world. 
%
%X: We address the challenge of Real2Sim2Real transfer for tasks involving elastic objects by performing accurate system identification using only a single interaction.\\ 
%Y: The primary bottleneck in the pipeline is the accurate Real2Sim transfer of physical parameters. Manual measurement is labor-intensive and requires specialized equipment, whereas conventional automatic system identification methods often lack the precision required for sensitive elastic dynamics or fail to capture the distinct elastic properties of the material.\\
%Z: To overcome these limitations, we propose a gradient-free hybrid optimization framework using a physics-aware loss function. We demonstrate the method’s efficacy through the successful Real2Sim2Real transfer of an RL policy for the "Slingshot" task.\\
\end{abstract}
\section{Introduction}
% Motivation
Elastic objects (EO) exhibit significant reversible deformation and enable a wide range of real-world applications, including manufacturing, healthcare, and everyday human-object interactions. Their behavior introduces high-dimensional deformation dynamics with entropic elasticity, where large strains lead to highly nonlinear responses~\cite{treloar1975physics}.
%small variations in material properties produce significantly different responses.
These characteristics make modeling, estimation, and control fundamentally challenging. Despite their practical importance, researchers have comparatively underexplored elastic object manipulation (EOM) relative to rigid or other deformable object manipulation (DOM). Developing robust manipulation strategies for elastic objects, therefore, remains a critical research objective.

\begin{figure}[t]
    \centering
    \includegraphics[width=\columnwidth]{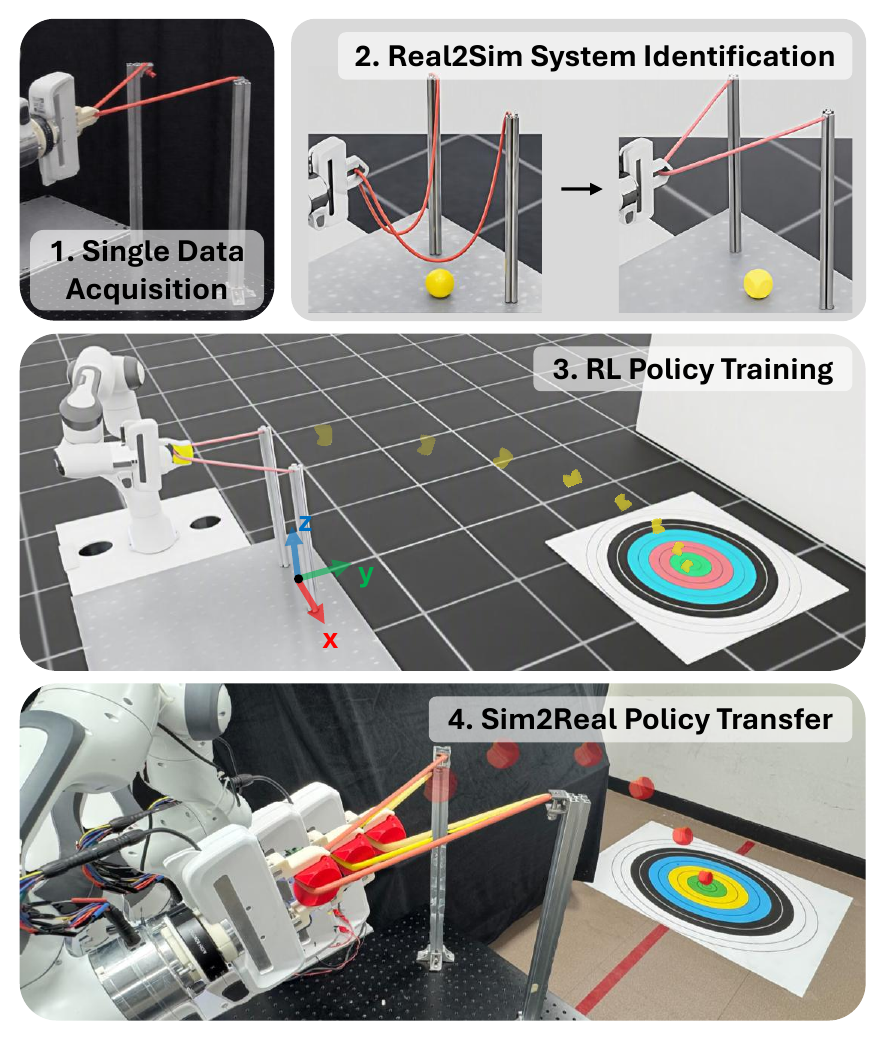} 
    \caption{An overview of the Sling2Sim2Real steps. After collecting a single non-destructive interaction with an unforeseen sling, the proposed method identifies its elastic parameters to construct a fine-fidelity simulator, enabling efficient slingshot policy learning for Sim2Real transfer. The learned policy then allows the manipulator to accurately hit the target with the slingshot.    
    %The bottom image shows that the robot needs to execute different actions to hit the target for different rubber bands. 
    }
    %\vspace{-2em}
    \label{fig:conceptualFigure}
\end{figure}

% Problem and Challenge
We aim to address EOM by leveraging the increasingly realistic physical fidelity of modern physics simulators. In this context, Real2Sim2Real has emerged as a representative paradigm for policy learning in rigid-body manipulation, particularly for unforeseen real-world objects~\cite{real2sim2real}. We investigate this problem in a slingshot manipulation environment, as shown in Fig.~\ref{fig:conceptualFigure}, involving rubber bands that exhibit diverse yet visually indistinguishable elastic properties (e.g., \textit{Young's modulus} and \textit{damping}). However, Real2Sim transfer in EOM remains non-trivial, as the visually indistinguishable nature of elastic properties makes it difficult to construct physically consistent simulated counterparts, thereby introducing a significant sim-to-real gap. Consequently, effective in-situ system identification (SI) of real-world elastic properties becomes essential for reliable policy learning and transfer.

% Previous Work 1 => Real2Sim (SI)를 통한 Real2Sim2Real
Prior Real2Sim2Real methods primarily reduce the sim-to-real gap through data-driven parameter inference and physics-consistent simulation modeling to enable efficient policy learning and transfer~\cite{real2sim2real}. However, these approaches typically require extensive real-world interaction data, assume sufficiently informative observations, and exhibit limited robustness when applied to elastic objects with visually indistinguishable and highly nonlinear physical properties. Limited interaction opportunities further impair SI, and resulting multimodal parameter landscape makes it difficult to identify an accurate parameter set. Therefore, we require an efficient but task-applicable SI method.

%data driven R2S => require huge data => more data on EO => need data efficient method 

%Another main challenge of this task is the large parameter space over typical SI targets (i.e., \textit{Young's modulus} and \textit{Poisson ratio}). The elasticity parameters (e.g., \textit{damping}) produce highly nonconvex and multimodal parameter landscape, which makes naïve local optimization unreliable. Despite their strong correlation
%\da{interaction requirement를 줄일 수 있는 노력이 필요함을 마지막에 강조 필요 => task-specific one-shot SI로 연결}

%기존 identification 방법론과 그 한계 => real-to-sim gap =>simulation 을 통한 synchronization

% Previous Work 2 => Real2Sim 
%Strong correlations between conventional deformable properties and damping parameters introduce a highly nonconvex and multimodal optimization landscape, 

% damping
% covariance-informed
% multi-start

% Solution
We propose Sling2Sim2Real, a one-shot Real2Sim2Real policy-learning framework for slingshot manipulation by leveraging covariance-informed multi-start optimization. Our framework consists of two stages: (1) real-to-sim SI given single interaction data and (2) sim-to-real policy learning in calibrated physics simulator; We collect a single, non-destructive visuo-haptic demonstration without launching the projectile. To enable effective SI, we capture both geometric and force responses to characterize elasticity. We then estimate correlations among five elasticity parameters and exploit this covariance structure to guide task-specific optimization. In particular, we perform global exploration using differential evolution (DE)~\cite{DE}, and subsequently refine multiple promising regions via multi-start covariance matrix adaptation evolution strategy (CMA-ES)~\cite{CMA_ES}. Finally, we train a reinforcement learning (RL) policy in the calibrated simulator and directly transfer it to the real system without additional adaptation.

We evaluate the proposed method against state-of-the-art baselines in a slingshot task involving diverse, unforeseen elastic bands in real-world settings. We implement Sling2Sim2Real on NVIDIA Isaac Sim~\cite{isaacsim}, identifying five parameters---including \textit{Young's modulus} and \textit{damping}---that critically govern rubber-like elastic material behavior but remain underexplored in prior real-to-sim SI studies. Our results demonstrate successful zero-shot sim-to-real policy transfer using a Franka Emika Panda arm. Performing ablation studies, we further verify the effectiveness of the hierarchical optimization architecture and covariance utilization.

\section{Related Works}

%\subsection{Deformable / Elastic Object Manipulation}
% DOM => EOM => ?
Early DOM methods drive DOs toward desired configurations using analytical models~\cite{jimenez2012survey} or task-specific heuristics~\cite{jia2018manipulating} for transition estimation. However, these approaches often suffer from limited robustness and poor generalization. Recent studies increasingly adopt data-driven algorithms that learn manipulation policies from sensory observations, including RGB images~\cite{matas2018sim} or point-cloud observations~\cite{Ze2024DP3}, at the cost of extensive data collection or exploration. Elasticity introduces nonlinear and highly sensitive deformation dynamics, further increasing the complexity of EOM and amplifying data requirements for stable policy learning~\cite{song2025implicit}. Consequently, researchers turn to simulation-based learning to reduce the cost of large-scale real-world data acquisition~\cite{lin2021softgym}. However, a number of studies assume known physical parameters or simplified simulation settings~\cite{song2025implicit}, leaving simulation fidelity and sim-to-real transfer remain fundamental challenges. We address the real-to-sim gap using one-shot system identification.

%. Despite the advantage of simulations, the sim-to-real domain gap limits transferring the learned policy to the real world. 

%INR-DOM enables an RL policy to capture the underlying representation and dynamics using implicit SDF, but the simulation and real world gap remain critical bottlenecks requiring known physical parameters or simplified simulation settings~\cite{song2025implicit}. 
%EOM works investigate rope, cloth, and soft-body manipulation~\cite{}, yet many methods assume known physical parameters or simplified simulation settings. Consequently, 

%\subsection{System Identification}
% Classic SI => Learning/Differential SI => SI in Robotics? 필요성? 한계? 
System identification plays a critical role in DOM ensuring physically consistent simulation~\cite{physTwin}, reliable state estimation~\cite{floren2024identification}, and robust policy transfer~\cite{kamaras2025distributional}. To construct reliable simulations, studies calibrate deformable properties, such as Young's modulus, to reduce the sim-to-real gap and reproduce contact behaviors; Early approaches estimate elasticity parameters from deformation observations using analytical formulations, including Hooke's law combined with finite element method~\cite{simulTrackingNSIofDO}. Recent approaches instead optimize simulation parameters by minimizing geometric, visual, or topological discrepancies between simulated and real-world objects using point observations~\cite{accurateSIofDLO, physTwin, paramLearning4packages} or images~\cite{bayesianTreatment}. To improve optimization efficiency, researchers increasingly adopt differentiable simulation frameworks~\cite{diffCloud, embodiedMpm, real2simDOM4TissueSurgery}. However, most methods primarily estimate soft-body parameters such as Young's modulus and Poisson ratio~\cite{diffCloud, physTwin, embodiedMpm, real2simDOM4TissueSurgery, bayesianTreatment, accurateSIofDLO, gendom}, while the stability and resilience of elastic object behavior in simulation strongly depend on damping effects. In contrast, we focus on the estimation Young's modulus and damping. Further, prior approaches predominantly rely on visual observations, despite the fact that haptic feedback provides direct and complementary cues for elasticity estimation~\cite{simulTrackingNSIofDO}, particularly when deformation or motion becomes visually ambiguous or occluded. 

%In addition to the visual difference, Sengupta et al. measures deformation forces to estimate elasticity parameters~\cite{simulTrackingNSIofDO}. 
%To the best of our knowledge, our work is the first work that jointly estimate conventional deformable properties with damping for elastic object SI. 
%System identification is a key factor to build an interactive digital twin for DOM in simulation by minimizing sim-to-real gap and compensating contact behaviors; otherwise, simulation policy fails to be deployed in real world though simulation provides efficient and stable learning.  

%\subsection{Real-to-Sim and Sim-to-Real Approaches}
% R2S2R => DO/EO 관련 한계 => 우리 연구 필요성
Despite precise real-to-sim calibration, sim-to-real policy learning often still requires extensive real-world interactions, as small modeling errors may induce significant policy degradation during transfer~\cite{zhao2020sim}. Instead, Aslam et al. perform RL directly in the real world using tactile observations, thereby bypassing simulation inaccuracies~\cite{dartbot}. Song et al. adopt a hybrid strategy that learns a latent representation in simulation while training the manipulation policy in the real world~\cite{song2025implicit}. In contrast, our work reduces modeling errors by simplifying the target EOM task as a one-step decision process, which reduces sensitivity to imperfect physics modeling and enables robust zero-shot deployment. 
\section{Methodology}
%Given a rubber sling with unknown elastic properties, the objective is to identify the underlying elastic parameters through system identification, and leverage the identified model to generate a control policy $\pi$ that stretches and releases the rubber thereby allowing the projectile to reach a specified target position $g$. 

\begin{figure*}[t]
    \centering
    % Make the figure 1.1 times the width of the text area
    % The makebox centers it, allowing it to spill into margins
    \makebox[\textwidth][c]{\includegraphics[width=\textwidth]{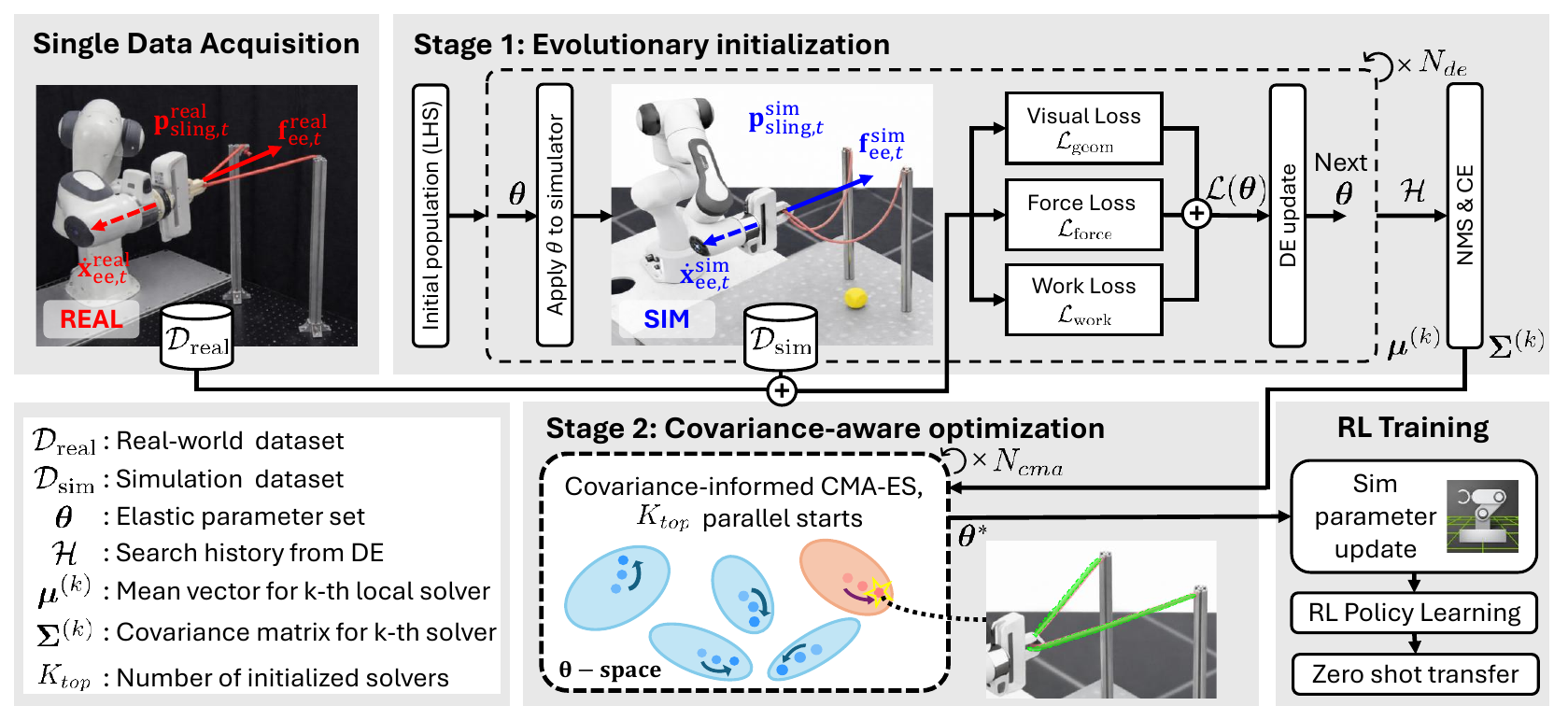}}
    \caption{An overview of the Sling2Sim2Real framework that performs one-shot Real2Sim system identification (SI) and Sim2Real policy transfer. We first record an interaction episode $\mathcal{D}_\text{real}$ by stretching and releasing the sling without a projectile. The Real2Sim SI then estimates elastic parameters, where differential evolution (DE) provides covariance-informed multi-start initialization to CMA-ES solvers. Resulting the high-fidelity simulator, the Sim2Real policy learner finally transfers a slingshot policy to the real-world platform. 
    } 
    \label{fig:overall_architecture}
\end{figure*}

%%%%%%%%%%%%%%%%%%%%%%%%%%%%%%%%%%%%%%%%%%%%%%%%%%%%
\subsection{Overall Framework}
We introduce Sling2Sim2Real, a two-stage policy learning framework that performs one-shot real-to-sim SI and then learns an optimal slingshot policy $\pi$ deployable in the real environment without real-world trial-and-error (see Fig.~\ref{fig:overall_architecture}). 

The first stage, \textbf{One-shot SI (Real2Sim)}, constructs a slingshot physics simulator in Isaac Sim through a real-to-sim pipeline. As the elastic properties of the sling are not known a priori, our pipeline calibrates elasticity-relevant simulation parameters from limited real-world interactions (i.e., controlled pulling) without projectile launches. We particularly calibrate a parameter set, $\bm{\theta}=\; $[\textit{Young's modulus}, \textit{Poisson's ratio}, \textit{elasticity damping}, \textit{dynamic friction}, \textit{damping scale}], by cycling between generating simulation data and updating the model parameters via numerical optimization. To effectively explore the complex parameter space, we introduce a hierarchical optimization framework of DE for global exploration and CMA-ES for parallelized local exploitations. 

The second stage, \textbf{Policy Learning (Sim2Real)}, then finds a real-world policy $\pi$ learning in the calibrated simulator using RL. In this work, we formulate the slingshot problem as a one-step Markov decision process (MDP) with random target positions. We deploy the learned policy $\pi^*$ directly to the real slingshot system without additional real-world policy refinement. 

%%%%%%%%%%%%%%%%%%%%%%%%%%%%%%%%%%%%%%%%%%%%%%%%%%%%
\subsection{One-shot System Identification (Real2Sim)}
We formulate SI as a black-box parameter estimation problem that minimizes the discrepancy between observed real single interaction data $\mathcal{D}_{\text{real}}$ and simulated responses $\mathcal{D}_{\text{sim}}$. We design a three-step Real2Sim identification pipeline that progressively improves parameter search robustness and convergence stability through structured single data acquisition, evolutionary initialization, and parallelized covariance-aware optimization (see Algorithm~\ref{alg:hybrid_sysid}).

\begin{algorithm}[h!]
\caption{Covariance-informed hierarchical system identification}
\label{alg:hybrid_sysid}
\begin{algorithmic}[1]
    % Expanded Require/Ensure to briefly label the inputs/outputs
    \Require $\mathcal{D}_{\text{real}}$ (Real-world dataset), $K_\text{top}$ (Number of initialized solvers), $K_\text{NN}$ (Neighbors used for covariance), $N_{\text{cma}}$, $N_{\text{de}}$ (Max. iterations for DE, CMA-ES)
    \Ensure $\bm{\theta}^*$ (Final optimized system parameters)

    % Added an unnumbered Definitions block for self-containment
    \item[\textbf{Definitions:}]
    \Statex $\mathcal{H}$: Search history from DE
    \Statex $\bm{\mu}^{(k)}$: Mean parameter vector for the $k$-th local solver
    \Statex $\mathbf{\Sigma}^{(k)}$: Covariance matrix for the $k$-th solver 
    \Statex $\mathbf{L}^{(k)}$: Cholesky decomposition factor of $\mathbf{\Sigma}^{(k)}$
    \Statex $\mathbf{z}^{(k)}$: Standard normal distribution samples
    \Statex $\mathcal{L}(\bm{\theta})$: Loss function evaluating parameters against $\mathcal{D}_{\text{real}}$
    \Statex % Blank line for spacing

    \State \textbf{// Stage 1: Evolutionary initialization}
    \State $\mathcal{H} \leftarrow \text{DE.Run}(\mathcal{D}_{\text{real}}, N_{de})$ %\Comment{Run DE and obtain history $\mathcal{H}$}
    \State $\{\bm{\mu}^{(k)}\}_{k=1}^{K_\text{top}} \leftarrow \text{NMS}(\mathcal{H}, K_\text{top})$ %\Comment{Extract centers}
    \State $\{\mathbf{\Sigma}^{(k)}\}_{k=1}^{K_\text{top}} \leftarrow \text{Covariance}(\mathcal{H},  \{\bm{\mu}^{(k)}\}_{k=1}^{K_\text{top}}, K_{NN}) $ %\Comment{Compute empirical covariance}
    \Statex % Blank line for spacing
        
    \State \textbf{// Stage 2: Covariance-aware optimization}
    \For{$k = 1 \dots K_{\text{top}}$}  
    \State $\mathbf{L}^{(k)} \leftarrow \text{Cholesky}(\mathbf{\Sigma}^{(k)})$        
    \State $\text{CMA}^{(k)} \leftarrow \text{CMA.Initialize}(\mathbf{z}_0 = \mathbf{0}, \sigma_0 = 1.0)$
    \EndFor
    \Statex % Blank line for spacing
    
    \For{$i = 1 \dots N_{\text{cma}}$}
        \State $\mathbf{z}^{(k)} = \text{CMA}^{(k)}\text{.Sample}() \quad \forall k$
        \State $\bm{\theta}^{(k)} \leftarrow \bm{\mu}^{(k)} + \mathbf{L}^{(k)} \cdot \mathbf{z}^{(k)} \quad \forall k$
        \State $\mathcal{L}_{\text{batch}} \leftarrow \text{EvaluateBatch}(\bigcup_{k=1}^K \bm{\theta}^{(k)}, \mathcal{D}_{\text{real}})$ %\Comment{Vectorized GPU Batch}
        \State $\text{CMA}^{(k)}\text{.Update}(\mathcal{L}_{\text{batch}}^{(k)}) \quad \forall k$
    \EndFor
    \Statex % Blank line for spacing

    \State \Return $\bm{\theta}^* \leftarrow \arg\min_{\bm{\theta} \in \mathcal{H}} \mathcal{L}(\bm{\theta})$
    %\vspace{-1em}
\end{algorithmic}
\end{algorithm}

\subsubsection{Single data acquisition}
We record non-destructive real-world interaction data by executing a sequence of primitive actions that grasps, stretches, and releases the sling without a projectile. In this work, we use a Franka Emika Panda, with an external RGB-D camera, that grasps the sling at its bottom point, pulls it by \qty{40}{\cm} from the plane of the U-shape fork along a predefined direction, and then releases it to excite the elastic dynamics. From the initial approach action to the position reset after release, we record synchronized robot proprioceptive signals and visual observations of sling deformation. Our measurements include end-effector velocity $\dot{\mathbf{x}}_{\text{ee},t}^{\text{real}}\in\mathbb{R}^3$, the interaction force $\mathbf{f}_{\text{ee},t}^{\text{real}}\in\mathbb{R}^3$ from the wrist-mounted force-torque sensor, and the sling point cloud $\mathbf{p}_{\text{sling},t}^{\text{real}}$ at each time step $t$. 
%\da{데이터 취득이 끝나는 시점이 어떻게 되나요? 2가지로 대답바랍니다. 1) 추후 saturation loss를 고려하지 않았을때와 2) 해당 loss를 추가하였을때.} \wk{데이터 취득은 robot action을 모두 수행 완료하였을때 종료됩니다. 이때 robot action이라는 것은 SI를 할 때 로봇이 고무를 잡아서 당기고 놓고 다시 home position으로 돌아가는 것까지 포함입니다. Saturation loss는 저번에 말씀드린 것처럼 실험해 본 결과 효과가 없거나, 결과에 오히려 악영향을 미쳐 사용하지 못할 것 같았습니다. 이때 saturation은 로봇이 고무를 release한 이후 고무의 움직임이 정지하는 것에 대한 saturation으로, robot action중 force가 saturate될 때 까지 기다리는 것과는 별개입니다.}
%\da{현재는, sling의 상태와 상관없이 로봇의 모션 종료까지네요? }

For accurate SI, we preprocess the raw observations to construct a single interaction episode $\mathcal{D}_\text{real}$:
\begin{align}
[(\dot{\mathbf{x}}_{\text{ee},0}^{\text{real}}, \mathbf{f}_{\text{ee},0}^{\text{real}}, \mathbf{p}_{\text{sling},0}^{\text{real}}), ..., (\dot{\mathbf{x}}_{\text{ee},T-1}^{\text{real}}, \mathbf{f}_{\text{ee},T-1}^{\text{real}}, \mathbf{p}_{\text{sling},T-1}^{\text{real}})],
\end{align}
which consists of $T$ time samples. To mitigate measurement noise, we apply a low-pass filter to the force signals using cutoff frequency \SI[per-mode=symbol] {0.224}{\radian\per\second}. To isolate the sling geometry, we segment the sling region in the RGB image using the Segment Anything Model~\cite{sam3}. We then project only the masked depth pixels into 3D space to generate the sling point cloud $\mathbf{p}_{\text{sling}}^{\text{real}}$. We further downsample the point cloud to $10^3$ points using farthest point sampling, followed by statistical and radius outlier removal. In this work, we set $T=300$ for \SI{9}{\second} and, for notational simplicity, use the same symbols to denote the preprocessed measurements.

\subsubsection{Evolutionary initialization}
We use DE to find $K$-number of distinct elastic parameter candidates $\{\bm{\theta}^{(1)}, \bm{\theta}^{(2)}, ..., \bm{\theta}^{(K)}\}$ for the subsequent parallel local optimizations. DE is a population-based evolutionary algorithm with global exploration, thereby holding a number of candidate solutions simultaneously. For the evolution, we use a fitness function represented as a composite loss:

\begin{align}
\mathcal{L}= \mathcal{L}_\text{geom} + \mathcal{L}_\text{force} + \mathcal{L}_\text{work},
\label{eq_loss}
\end{align}
where $\mathcal{L}_\text{geom}$ measures the visual deformation gap and the others represent physical transition gaps. 

% \da{마지막에 history기반? saturation time 관련 loss를 추가한다고 했던거 같은데 어디에? }
% \wk{아직 3가지 고무에 대해 한 cycle을 돌지 못해, 한 cycle을 모두 돌고, saturation-based DE history filtering을 사용했을 때의 결과가 유의미하다는 것이 입증되었을 때 해당 내용을 추가하려고 합니다.}
% \da{그렇게 하나씩 체크할 여유가 없어보이는데, 따로 돌릴수는 없나요?  }
% \wk{현재 사용 가능한 GPU를 최대한으로 끌어서 사용하고 있지만, 추가적으로 사용할 수 있는 GPU가 있는지 알아보도록 하겠습니다.}

The loss $\mathcal{L}_\text{geom}$ is to minimize the visual mismatch between real and simulated sling geometry due to uncalibrated parameters. To measure the mismatch, we use an aggregated Chamfer distances (CD) as the $\mathcal{L}_\text{geom}$, 
\begin{align}
    \mathcal{L}_{\text{geom}} &= \sum_{t=0}^{T-1} w_\text{geom} \cdot \text{CD}(\mathbf{p}_{\text{sling},t}^{\text{real}}, \mathbf{p}_{\text{sling},t}^{\text{sim}}) ,
    \label{eq_geom_loss}
\end{align}
where $w_\text{geom}$, $CD(,)$, and $\mathbf{p}_{\text{sling},t}^{\text{sim}}$  are a constant, a CD function between two point clouds, and a segmented sling point cloud from the simulator, respectively. 
%We particularly use a time-varying weight $w_t=\left( 1 + \exp\left( 0.5 (t - t_{\text{open}}) \right) \right)^{-1}$ assigning a higher weight until the robot releases the sling since the interaction forces and vibration discrepancies become the dominant factors for parameter calibration by opening the gripper at $t_{\text{open}}=270$.
%\wk{$t_{\text{open}}$ is the exact time step at which the command to open the gripper is transmitted. }
The remaining terms, $\mathcal{L}_\text{force}$ and $\mathcal{L}_\text{work}$, measure the force and work discrepancies, respectively. We define
\begin{align}
    % Point-wise force comparison
    \mathcal{L}_{\text{force}} = w_\text{force} \cdot &\sum_{t = 0}^{T-1} \| \mathbf{f}_{\text{ee},t}^{\text{real}} - \mathbf{f}_{\text{ee},t}^{\text{sim}} \|_2 , \\
    \mathcal{L}_{\text{work}} = w_\text{work} | &\frac{1}{2}\sum_{t=0}^{T-1} ( 
    \mathbf{f}_{\text{ee},t}^{\text{real}} \cdot \dot{\mathbf{x}}_{\text{ee},t}^{\text{real}} 
    + \mathbf{f}_{\text{ee},t+1}^{\text{real}} \cdot \dot{\mathbf{x}}_{\text{ee},t+1}^{\text{real}} \nonumber \\
    &- \mathbf{f}_{\text{ee},t}^{\text{sim}} \cdot \dot{\mathbf{x}}_{\text{ee},t}^{\text{sim}} 
    - \mathbf{f}_{\text{ee},t+1}^{\text{sim}} \cdot \dot{\mathbf{x}}_{\text{ee},t+1}^{\text{sim}} ) | ,
    \label{eq_interaction_losses}
\end{align}
where $w_\text{force}$ and $w_\text{work}$ are constants. The force-matching loss $\mathcal{L}_\text{force}$ directly calibrates the stiffness of the simulated elastic object by aligning the restoring force at each time step. However, this sample-wise comparison is vulnerable to noisy force signals and temporal misalignments between simulation and real data, resulting in significant loss increases. To improve robustness, we introduce the work-based loss $\mathcal{L}_\text{work}$, defined as the accumulated force-velocity product. This formulation enforces consistency in the mechanical work between simulated and real interactions, enabling calibration of both the overall stiffness scale and the dissipated energy governed by damping. We estimate the fitness value $\mathcal{L}(\bm{\theta})$ of the current parameters $\bm{\theta}$ during DE by generating single-interaction data $\mathcal{D}_{\text{sim}}$ at each iteration through simulation. Note that, in this work, we smooth the force and velocity signals via a median filter with a window size of $7$ before computing the losses. We generate initial population of parameters using latin hypercube sampling.  
%\da{$\mathcal{L}_\text{work}$ 만으로 충분한게 아닌가요? $\mathcal{L}_\text{force}$가 필요한 이유가 뭔가요?}
%\wk{Force의 직접적인 비교는 두 force plot 간의 차이를 가장 정확하게 파악할 수 있게 해줍니다. 하지만 force만을 사용하면 sensor noise 등에서 파생되는 noise에 너무 큰 영향을 받게 됩니다. 따라서 전체 interaction 동안의 에너지를 나타내는 work를 계산하여, noise의 영향을 mitigate하는 일종의 regularization term으로 사용합니다. 다만 work만 사용하면 곡선 아래의 넓이(에너지 총량)만 보게 되므로 물리적 거동의 세부적인 형태에 대한 ambiguity가 생길 수 있습니다. 결론적으로 force와 work를 모두 사용함으로써 정밀한 비교와 regularization 사이의 균형을 맞추기 위해 loss를 이와 같이 formulate했습니다.}\da{ablation에서 증명바랍니다.}

By retaining all candidate solutions generated during evolution in the history container $\mathcal{H}$, we extract the top-$K_\text{top}$ parameter regions and construct their corresponding center-covariance pairs $\{(\bm{\mu}^{(k)}, \mathbf{\Sigma}^{(k)})\}_{k=1}^{K_\text{top}}$. To identify distinct regions, we apply a distance-based non-maximum suppression (NMS) with suppression radius $\delta$. After sorting the DE solutions $[\bm{\theta}^{(1)}_{DE},  \bm{\theta}^{(2)}_{DE}, ... , ]$ in ascending order of loss, we iteratively keep a solution $\bm{\theta}^{(k)}_{DE}$ as a new region center $\bm{\mu}^{|\mathcal{G}|+1}$ if $\| \bm{\theta}^{(k)}_{DE} - \bm{\theta}^{(i)}_{DE} \|_2 > \delta$ and $\forall \bm{\theta}^{(i)}_{DE} \in \mathcal{G} $ where $\mathcal{G}$ denotes the set of already selected centers. We repeat this procedure until $|\mathcal{G}|=K_\text{top}$. For each selected center $\bm{\mu}^{(k)}\in\mathcal{G}$, we then estimate the covariance matrix $\mathbf{\Sigma}^{(k)}$ by constructing $K_\text{NN}$-nearest neighbor group in $\mathcal{H}$ and computing the empirical covariance over the associated neighborhood. In this work, we use $K_\text{top}=5$, $K_\text{NN}=15$, and $\delta=0.1$. We use these pairs as multi-start initialization distributions for subsequent parameter optimization in Sec.~\ref{ssec_cma_es}. 
%\da{실제 사용된 $K_\text{top}=?$ and $K_\text{NN}=?$ } \wk{$K_\text{top}=5$, $K_\text{NN}=15$ and $\delta=0.1$}

%we find a $K$-number of non-overlapping index groups $\mathcal{G}_{1:K}$ in order where $\mathcal{G}=\{ j\; |\; \| \bm{\theta}^{(j)}_{DE} - \mathbf{\mu}^{(k)} \|_2 \leq \delta, j \notin \mathcal{G}_i \}$, $\mu^{(k)}$ is the centroid of the $i$-th group solutions, and $\delta$ is a threshold. 

\subsubsection{Covariance-aware optimization}\label{ssec_cma_es}
Given the $K_\text{top}$ starting points, including $\{(\bm{\mu}^{(k)}, \mathbf{\Sigma}^{(k)})\}_{k=1}^{K_\text{top}}$, we run parallel CMA-ES solvers with the real episode $\mathcal{D}_\text{real}$ and the loss $\mathcal{L}$ in Eq.~(\ref{eq_loss}) to estimate the optimal parameter set $\bm{\theta}^*$. This complements the suboptimal global exploration of DE in the highly nonlinear elasticity parameter space under noisy observations. By leveraging the covariance matrices estimated from the diverse DE candidates, each CMA-ES instance initializes a locally structured search distribution, enabling efficient local optimization around promising regions. The parallel CMA-ES solvers iteratively update their candidate solutions adapting the search distribution, thereby capturing underlying parameter correlations and accelerating convergence toward consistent elastic properties. Note that we also compute the loss by running simulations for each solver at every iteration.
%\da{전체적으로 DE와 CMA-ES 중간에 시뮬레이션을 돌려서 비교한다는 내용이 빠져있음.} \wk{교수님 수정 완료하시면 추후에 optimization의 각 iteration 마다 sim에서 evaluation을 하고 $\mathcal{D}_{\text{sim}}$ 을 얻는다는 것을 설명하도록 하겠습니다.}

In detail, we transfer each start prior $(\bm{\mu}^{(k)}, \mathbf{\Sigma}^{(k)})$ to CMA-ES by re-parameterizing the search space through a Cholesky decomposition of the covariance, $\mathbf{\Sigma}^{(k)} = \mathbf{L}^{(k)} \mathbf{L}^{(k)T}$. Instead of sampling directly in the space, we draw candidates from an isotropic latent Gaussian $\mathbf{z} \sim \mathcal{N}(\mathbf{0}, \mathbf{I})$ and map them via $\bm{\theta}_{\text{init}} = \bm{\mu}^{(k)} + \mathbf{L}^{(k)} \mathbf{z}$ similar to \cite{deCmaHybrid}. This transformation reshapes the initial spherical search distribution into an anisotropic one aligned with the local curvature and parameter correlations of each promising region. By conditioning the $K_{\text{TOP}}$ parallel solvers with the covariance priors and selecting the best solution, we enable more efficient and accurate local optimization in this high-dimensional DO SI. %\da{covariance 를 CMA-ES로 연결하는 연구 reference?}
%\wk{\cite{deCmaHybrid}}

%%%%%%%%%%%%%%%%%%%%%%%%%%%%%%%%%%%%%%%%%%%%%%%%%%%%%%%%%%%%%
\subsection{Sim2Real Policy Learning}
After constructing a high-fidelity sling simulation, we formulate the slingshot task as a one-step MDP. Let $\Pi_{||}$ denote the plane of the U-shape fork, and let $\Pi_{\perp}$ denote the projectile plane that is perpendicular to $\Pi_{||}$ and passes through both the fork center and the target. We define a reference frame $\mathcal{O}$ at the midpoint between the two fork tips on $\Pi_{||}$, where the $y$-axis points from the fork center toward the target and $z$-axis aligns with the vertical direction. We then define an MDP tuple $<\mathcal{S}, \mathcal{A}, \mathcal{T}, R>$, where the elements are as follows
\begin{itemize}[leftmargin=*]
\item $\mathcal{S}$: a set of states describing the system configuration before and after the projectile launch. We define each state as $s_t=(\mathbf{x}_{\text{proj},t}, g)$, where $\mathbf{x}_{\text{proj},t}=(y_{p,t},z_{p,t} )\in\mathbb{R}^2$ represents the 2D projectile position $(y_{p,t},z_{p,t})$ within $\Pi_{\perp}$, and $g\in\mathbb{R}$ denotes the target displacement from the origin. We constrain the initial state $s_0\in\mathcal{S}$ to the configuration in which the projectile is attached to the sling. We define the terminal state $s_1$ as the configuration when the projectile contacts the target plane after launch, triggered by releasing the gripper.
%$s_t=(\Phi_{\text{sling},t}, \Phi_{\text{proj},t}, g)$, where $\Phi_\text{sling}$ denotes the sling state comprising its geometry and material properties, $\Phi_\text{proj}$ represents the projectile position and orientation, and $g$ indicates the target position. To simplify this problem into a single-shot, single-step problem, we assume that the initial state $s_0\in\mathcal{S}$ is when the projectile is placed on the sling, thereby $s_t=(\Phi_{\text{proj},t}, g)$.
\item $\mathcal{A}$: an action space where each action, denoted by $\mathbf{x}_\text{proj}\in\mathbb{R}^2$, defines the drawn position of the projectile. In this work, we align the projectile direction (i.e., elevation angle) with the plane formed by the stretched sling for stable launch,
\item $\mathcal{T}$: a transition model $\mathcal{T}:\mathcal{S}\times\mathcal{A}\rightarrow Pr(\mathcal{S})$ where $Pr(\mathcal{S})$ denotes the set of all probability distributions over $\mathcal{S}$. In this MDP, a transition occurs from the initial state to a terminal state, and
\item $R$: a reward function $R:\mathcal{S}\times\mathcal{A}\times\mathcal{S}\rightarrow\mathbb{R}$, which denotes a dartboard-style impact-location-based reward: $R = 5e^{-|{y_p}-g|}$.
%
% \begin{align}
%     R &= 
%     \begin{cases} 
%         4e^{-|y_{p}-g|} + 1.15, & \text{if } |y_{p}-g| \le \qty{30}{cm}  \\
%         3e^{-\frac{(y_{p}-g)^2}{2}} + 0.25, & \text{if } \qty{30}{cm} < |y_{p}-g| \le \qty{50}{cm} \\
%         2e^{-\frac{(y_{p}-g)^2}{4.5}}, & \text{otherwise},
%     \end{cases}\nonumber
% \end{align}
%
In addition, we assign a penalty of $1$ to unsuccessful launches (e.g., unexpected contact or fall). In this work, \SI{20}{\cm} denotes the radius of the target ``bullseye." 
\end{itemize}
We train the optimal policy by maximizing the expected impact score under the calibrated transition dynamics: $\pi^*(s)=\arg\max_{a\in\mathcal{A}}\mathbb{E}[R|s,a]$.

\section{Experiment Setup}\label{sec_setup}
Our experiments aim to answer two key questions: (1) Does our Sling2Sim2Real framework provide an accurate zero-shot RL policy for the slingshot task? (2) Further, do our proposed covariance-informed multi-start initialization and loss formulation improve the Real2Sim system identification? We perform quantitative evaluations through the slingshot task with a real robot.

\subsection{Slingshot Environment Setup}
We introduce a real-world slingshot platform and its replicated simulation environment for Real2Sim2Real evaluation. As shown in Fig.~\ref{fig:conceptualFigure}, the real-world setup consists of a 7 degree-of-freedom manipulator (Franka Emika Panda), a wrist-mounted $6$-axis force-torque (FT) sensor (Aidin Robotics AFT200), and an external RGB-D camera (Orbbec Femto Bolt). We mount the robot in front of a U-shape slingshot fork composed of two vertical rods, separated by \SI{27.5}{\cm}. We fix a rubber band with a resting length of \SI{52}{\cm} to the tips of two rods. Using the rubber sling, the manipulator launches a 3D-printed projectile with a mass of \SI{39}{g} toward a $1\unit{\meter}\times1\unit{\meter}$ archery target placed on the floor.
%each \SI{52}{\cm} in height and 

We then replicate the real-world setup in Isaac Sim by spawning an identical simulated robot and slingshot fork. For SI, we model the sling as a tetrahedral cylindrical mesh with a length of \SI{55}{\cm}, including the clamping region, and a diameter of \SI{1}{\cm}. We parameterize the model with five elasticity-related variables, initialized with simulator's default deformable parameter values. To achieve high-fidelity simulation, we run Isaac Sim with a simulation time step of $\frac{1}{200}\mathrm{s}$ and a solver iteration count $64$ on a workstation equipped with an AMD Ryzen Threadripper Pro 5975WS and NVIDIA RTX 6000 Ada GPUs. Due to real-time constraints, however, we conduct SI using pre-collected interaction data rather than online estimation. 
%: \textit{Young's modulus}$=5e^{7}$, \textit{Poisson's ratio}$=0.45$, \textit{damping}$=5e^{-3}$, \textit{dynamic friction}$=0.25$, \textit{damping scale}$=1.0$
% \da{고무의 지름이 각각 다른데, 미리 mesh를 이에맞게 준비해서 만들었나요? 아니면 동일한 mesh를 사용했나요?}
% \da{mesh 길이는?}
% \wk{고무의 지름이 다르지만, 그 차이가 2mm밖에 되지 않는 점을 고려하였을때 로봇과 interaction을 하는데에 있어서 영향을 주지 않고, 고무가 더 두꺼움으로 인해 생기는 차이는 물성에 implicit하게 들어날 것으로 생각하여 직경을 모두 1cm로 통일하였습니다. 고무 길이는 real에서는 52cm 인데, sim에서 attatchment를 생성하기 위해 알루미늄 프로필과 겹쳐지는 부분과, real에서 clamping을 했을때 생기는 offset를 측정하여 양 끝에서 그 차이가 1.5cm가 되는것을 확인하였고, 그 차이를 고려하여 sim  mesh는 55cm로 설정하였습니다.}
% \da{어떻게 simulated sling의 initial parameter를 셋팅했나요? } \wk{USD asset에서는 IsaacSim의 비정형체 물성 기본값인 Youngs's Modulus = 5e7, Poisson's Ratio = 0.45, Damping = 5e-3, Dynamic Friction = 0.25, Damping Scale = 1.0으로 설정하였습니다. 하지만 System ID를 할때는 scene이 reset될 때 마다 물성이 바로 update되고, 초기 상태에도 optimizer의 initialization값을 사용하기 때문에 USD에서 mesh의 물성값은 영향을 주지 않을 것 같습니다.}
% \da{DE를 돌릴때 search의 초기값은 USD에서의 값에서 시작하는거 아닌가요?}
% \wk{아닙니다. USD의 물성치는 전혀 양향을 주지 않고, Optimization의 initial value가 첫 iteration부터 적용되어 optimization을 진행하고 있습니다.}
% \da{DE의 초기치가 얼마인가요? 그게 초기값이라고 봐야하는데? } \wk{맞습니다. 하지만 하나의 초기값이 있는것이 아니고, population안에 있는 각각의 agent가 서로 다른 값으로 initialize되고 있습니다. Initial population 을 sampling하는 방식은 latin hypercube 입니다.}
%\da{solver iteration count는?} \wk{64로 설정하였습니다.}
% \da{SI simulation workstation spec은?} \wk{CPU: AMD Ryzen Threadripper Pro 5975WX / RAM: 256GB / GPU: NVIDIA RTX 6000 Ada Generation 4개}\da{RTX 6000 4개를 다 사용하는 것인가요?}

To record real-world calibration data, we command the robot to perform predefined non-destructive interaction behaviors, including \textit{holding}, \textit{stretching}, and \textit{releasing} the uncalibrated rubber sling. We generate cubic-spline trajectories through predefined waypoints to ensure smooth and consistent interaction, inserting a short pause before release until the tension saturates. During data acquisition, we sample the wrist-mounted FT sensor at \qty{1}{kHz} and capture the external camera at $30$ FPS. For synchronized logging, we downsample the FT measurement to \qty{30}{\hertz} to align them with the visual observations. 
% \da{고무의 internal tension이 saturate될때까지 로봇이 고무를 잡고 있다고 하였는데, 지난 비디오에서는 딱히 그 장면을 보는 것은 어려웠습니다. saturation이 되는 것을 어떻게 측정하였나요?}

% \wk{이번에 새롭게 데이터를 측정할 때 saturation phase를 추가하여 로봇이 release position에서 고무를 잡고 정지하는 단계를 추가했습니다. Saturation phase를 충분히 길게 하여 force peak가 plateu하도록 하는 것이 목표이고, 실제 F-t graph에서도 force가 plateau하는 것을 확인하였습니다. 또한 해당 phase를 추가하고, interpolation step수를 조정하여 현재는 $T=300$을 사용하고 있습니다. 수정된 사항 바로 말씀 못 드린 점 사과드립니다.}

\begin{figure}[t]
    \centering
    % First image
    \includegraphics[height=3.45cm]{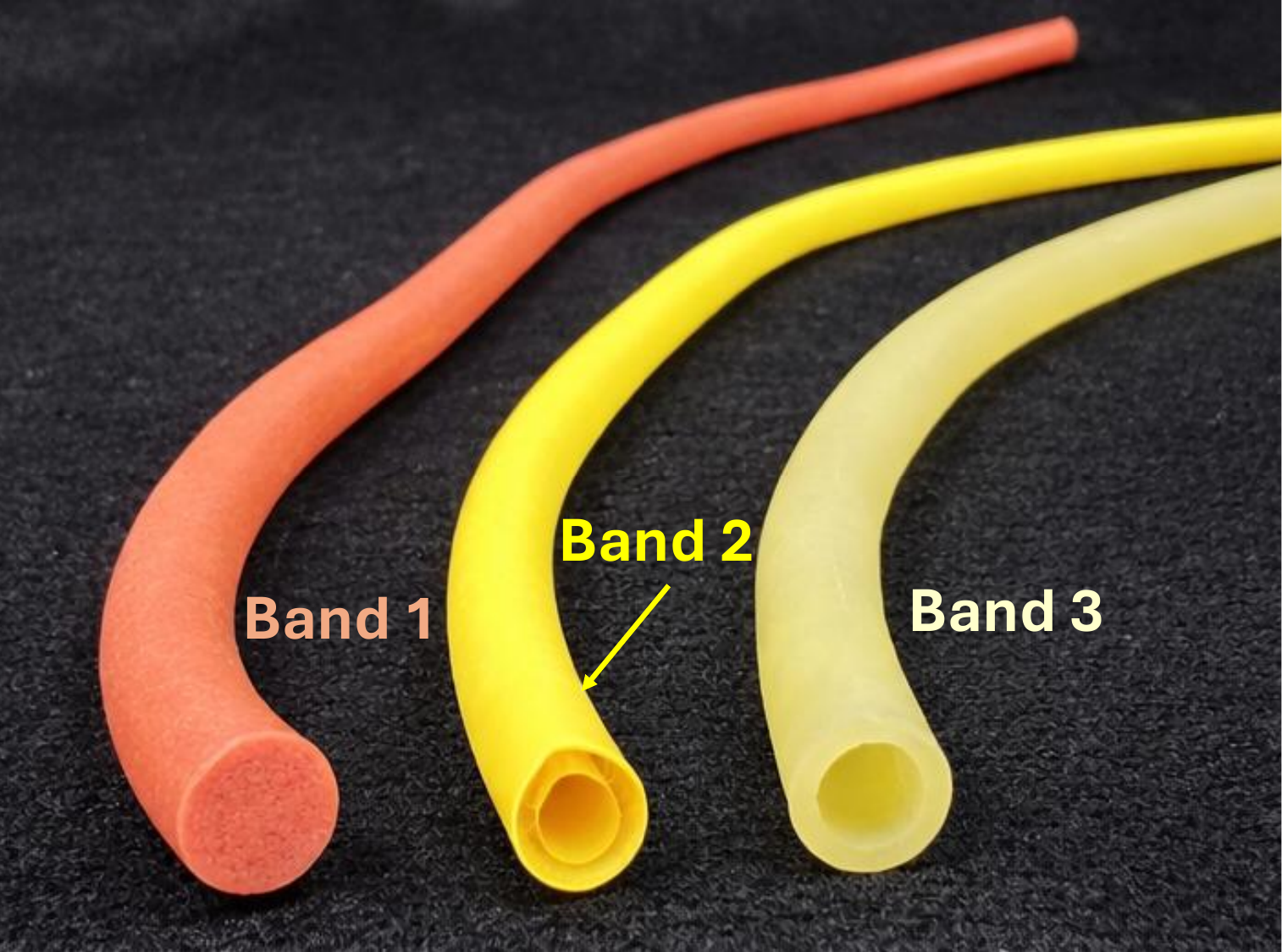}%
    % Second image
    \includegraphics[height=3.45cm]{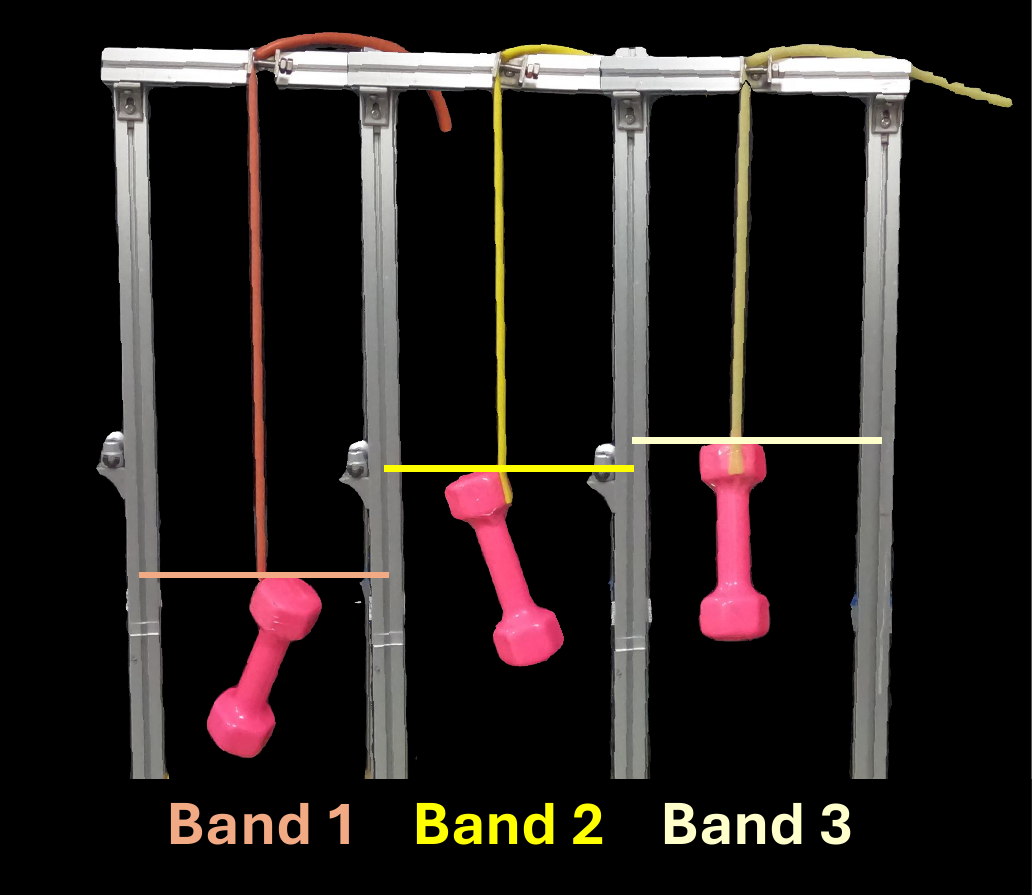}
    
    \caption{Three elastic bands used in the quantitative evaluation. The right image shows the elongation of each rubber band when a \qty{1.5}{kg} dumbbell is attached. Note that we clamp each rubber band at half of its full length.}
    % \wk{Visual적으로 많은 차이는 나지 않는 것 같습니다...}
    % \da{꼭 fork에 달아서 보여주지말고 한줄로 길게달아서 찍어보세요. 그럼 차이가 2배가 되겠죠.}
    \label{fig:rubber_bands}
    %\vspace{-1.5em}
\end{figure}

\subsection{Quantitative Evaluation}
We statistically evaluate the Real2Sim2Real SI performance against state-of-the-art baselines under the variation of slings and slingshot target locations. In detail, we use three rubber slings with a resting length of \SI{52}{\cm} (see Fig.~\ref{fig:rubber_bands}):
\begin{itemize}[leftmargin=*]
\item \textbf{Band 1}: silicon rope (diameter \SI{10}{\mm}, mass \qty{23}{g}), exerting a maximum force of \SI{12.18}{\newton} when pulled in SI;
\item \textbf{Band 2}: natural rubber hose (diameter \SI{8}{\mm}, mass \qty{17}{g}), exerting a maximum force of \SI{14.25}{\newton}; and
\item \textbf{Band 3}: latex sport band (diameter \SI{10}{\mm}, mass \qty{21}{g}), exerting a maximum force of \SI{16.7}{\newton}.
\end{itemize}
%\wk{각 Band의 질량 정보 Band 1: \qty{23}{g} / Band 2: \qty{17}{g} / Band 3: \qty{21}{g}를 추가하면 좋을 것 같습니다}
%
To evaluate the SI performance, we vary the target displacement as $g=\{172.5\unit{\cm}, 192.5\unit{\cm}, 212.5\unit{\cm}\}$. For each band-target pair, we conduct five launch trials per method. For comparison, we consider five baseline methods:
\begin{itemize}[leftmargin=*]
\item Real2Sim2Real~\cite{real2sim2real}: a simulation-based SI and policy training method for table-top deformable objects. This method relies solely on DE-based global parameter search and calibrates $[\textit{Young's modulus}, \textit{dynamic friction}]$;
%does not consider \textit{damping}-related elastic properties;
\item Package SI~\cite{paramLearning4packages}: a hyperelastic deformable package modeling method that optimizes $[\textit{Young's modulus},\textit{damping}]$ via Bayesian optimization by minimizing visually observed geometric discrepancies;
\item $\text{DE}_{\mathcal{L}}$: a simple baseline that calibrates the five elastic parameters via DE by minimizing the composite deformation gap $\mathcal{L}$ in Eq.~(\ref{eq_loss});
\item $\text{BO}_{\mathcal{L}}$: a simple baseline that calibrates the five elastic parameters via Bayesian optimization by minimizing $\mathcal{L}$ in Eq.~(\ref{eq_loss}); and
\item $\text{Adam}_{\mathcal{L}}$: a simple baseline that calibrates the five elastic parameters via Adam optimization by minimizing $\mathcal{L}$ in Eq.~(\ref{eq_loss}). We compute simulation gradients using finite-difference approximation, similar to DiffCloud~\cite{diffCloud}.
%Real2Sim2Real~\cite{real2sim2real}: a simulation-based SI and policy training method for table-top deformable objects. Unlike ours, this method relies solely on DE-based global parameter search and does not consider \textit{damping}-related elastic properties;
%Deformable package SI~\cite{paramLearning4packages}: a hyperelastic deformable package modeling method that optimizes three parameters, including \textit{damping}, via Bayesian optimization by minimizing visually observed geometric discrepancies; and 
%Modified DiffCloud~\cite{diffCloud}: a differentiable simulation-based SI method that aligns \textit{stiffness}, \textit{mass}, and \textit{friction}. In this work, we use finite difference due to the 
\end{itemize}

By integrating the identified parameter with Isaac Sim, we train an RL policy using proximal policy optimization (PPO) implemented via the RSL-RL library \cite{rsl_rl} for each method. We parameterize both the actor and critic networks with hidden dimensions of $[256, 128, 64]$ and ELU activations. To assess performance, we measure the projectile landing errors.

\section{Evaluation}

\noindent\textbf{Benchmark studies.}
% Baseline comparison
We first statistically evaluate the combination of Real2Sim SI and Sim2Real policy transfer performance of the proposed Sling2Sim2Real method by measuring the landing distance errors between the target and the point of impact.  
Table~\ref{tab:rl_eval} shows that Sling2Sim2Real consistently outperforms all baseline methods, achieving the lowest landing distance errors across all rubber band types, as shown in the rightmost column. The stiffer the band, the higher the landing distance error becomes. However, Sling2Sim2Real is able to successfully launch the projectile in the target area with minimum landing distance errors. Baselines occasionally result in `N/A,' indicating that the method either fails to launch the projectile due to excessive stretching force or launches it outside the target area. As shown in Table~\ref{si_loss}, two baselines estimate higher \textit{damping} scales, which likely reduce the returning kinetic energy and require greater stretching to launch the projectile. In addition, the error magnitudes vary significantly with the target distance $g$, as a small displacement of \qty{1}{cm} in the launch pullback position can result in a landing deviation of up to \qty{30}{cm} in our slingshot task.

%indicating the highly nonlinear parameter space. Throughout our experiment, we also found a minor displacement of \qty{1}{cm} in the launch pullback position results in a landing deviation of \qty{30}{cm}, thereby the optimization error in the 

%, making successful policy transfer highly dependent on the precision of SI.

%\da{수치 일관성 문제 언급 필요.}

%The failure represents the large Sim2Real gap in policy learning caused by the large Real2Sim SI errors. 

Table~\ref{si_loss} demonstrates the effectiveness of the proposed composite loss $\mathcal{L}$ for calibrating high-fidelity elastic simulation parameters. Lower landing distance errors correspond to lower composite loss. In particular, Sling2Sim2Real achieves the lowest calibration loss $\mathcal{L}$ for all band types, with a maximum reduction of $72.5\%$ compared to the next best baseline, Package SI~\cite{paramLearning4packages}. Although Package SI calibrates \textit{Young's modulus} and \textit{damping}, two dominant factors governing elasticity behavior, its geometry-based loss fails to capture stored elastic energy under high-damping materials (i.e., Band 1 and 2). As a result, Package SI produces higher losses and larger landing distance errors than our Sling2Sim2Real in most cases. In contrast, Real2Sim2Real~\cite{real2sim2real}, which calibrates only \textit{Young's modulus} and \textit{dynamic friction} with higher losses, fails to identify transferable parameters for effective Sim2Real policy learning. 

%while the method achieves similar or better performance in the case of the farthest target that requires high recovery

% \da{테이블을 보면 수치의 경향성이 약간 일관적이지 않습니다. 특히 g=1925일때 수치가 매우 떨어지는데 이유가 뭘까요? }
% \wk{우선 Single data acquisition단계에서 수행하는 action과 RL policy에서 나온 action이 비슷(in-distribution)할때 target을 잘 맞추는 가능성을 생각해보았는데요, 만약 그것이 문제였다면 soft는 $g = 172.5$거리를 가장 잘 맞췄어야 할 것인데, 그런 경향성이 보이지 않고 있습니다. 또한, 모든 고무에 대해 항상  $192.5$ 거리를 가장 잘 맞추는 것은 아니고 특정 패턴이 존재하지 않는것으로 보아, SI로 찾은 물성과, IsaacSim자체의 sim2real gap이 복합적으로 작용하여 나타나는 현상인 것 같습니다.}\da{음.. 간단히.. 아직 모르겠네요.}

%Given the highest stiffness of `Band 3,' the proposed method is only able to launch the projectile in the measurable target area while the other two baselines fails to launch the projectile or launches it outside the target area in all cases of target distances. Given comparably soft two bands, the proposed method outperforms baselines where. 

\begin{figure*}[t] % Use figure* for full width in two-column documents
    \centering
    \includegraphics[width=\textwidth]{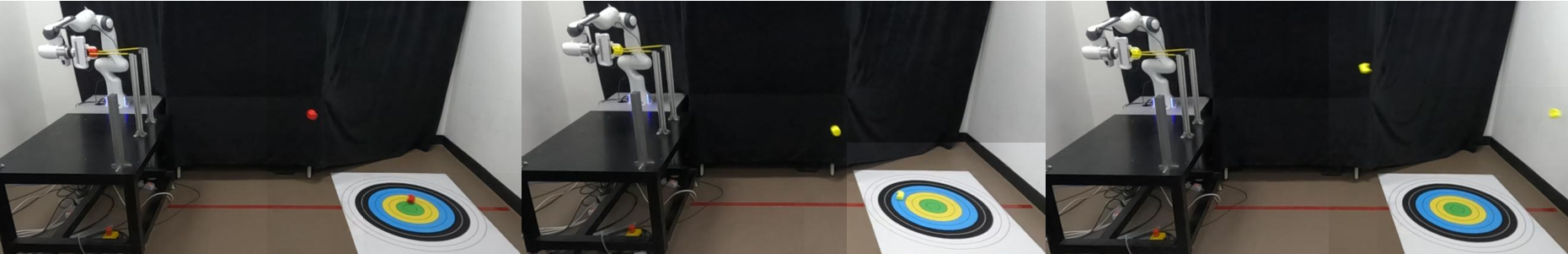}
    \caption{Real-world demonstrations of Sim2Real results. The overlaid traces show the projectile trajectories after launch. From left to right, the images correspond to Sling2Sim2Real, Package SI \cite{paramLearning4packages}, and Real2Sim2Real \cite{real2sim2real}}.
    \label{fig:rl_comparison_visual}
\end{figure*}

\begin{table}[t]
    \centering
    \caption{Comparison of projectile landing accuracy (distance error) in the slingshot environment across three target distances from the origin. `N/A' indicates that the resulting policy either fails to launch the projectile or launches it outside the target area, making the distance error unmeasurable.}
    \label{tab:rl_eval}
    
    \resizebox{\columnwidth}{!}{%
    \begin{tabular}{llcccc} 
        \toprule
        \multirow{2}{*}{\textbf{Material}} & \multirow{2}{*}{\textbf{Method}} & \multicolumn{4}{c}{\textbf{Landing Distance Error (\unit{\cm}) $\downarrow$}}\\
        \cmidrule(lr){3-6}
        & & $g=172.5$ & $g=192.5$ & $g=212.5$ & All Targets\\
        \midrule
        
        \multirow{3}{*}{\shortstack{\textbf{Band 1}\\\textbf{(Soft)}}}
        & Real2Sim2Real~\cite{real2sim2real}          & N/A & N/A & N/A & N/A \\
        & Package SI~\cite{paramLearning4packages} & $13.00 \pm 1.87$ & $1.50 \pm 1.22$ & $\mathbf{9.50 \pm 2.50}$ & $8.00 \pm 1.94$ \\
        & Sling2Sim2Real                 & $\mathbf{11.00 \pm 2.00}$  & $\mathbf{1.00 \pm 1.22}$  & $\mathbf{9.50 \pm 4.30}$ & $\mathbf{7.17 \pm 2.68}$ \\
        \midrule

        \multirow{3}{*}{\shortstack{\textbf{Band 2}\\\textbf{(Medium)}}}
        & Real2Sim2Real~\cite{real2sim2real}          & N/A & N/A & N/A & N/A \\
        & Package SI~\cite{paramLearning4packages} & $60.00 \pm 2.74$ & $34.50 \pm 3.32$ & $\mathbf{11.00 \pm 6.44}$ & $35.17 \pm 4.58$ \\
        & Sling2Sim2Real                 & $\mathbf{15.50 \pm 1.87}$  & $\mathbf{3.50 \pm 2.00}$  & $23.50 \pm 2.55$ & $\mathbf{14.17 \pm 2.16}$ \\
        \midrule

        \multirow{3}{*}{\shortstack{\textbf{Band 3}\\\textbf{(Stiff)}}}
        & Real2Sim2Real~\cite{real2sim2real}          & N/A & N/A & N/A & N/A \\
        & Package SI~\cite{paramLearning4packages} & N/A & N/A & N/A & N/A \\
        & Sling2Sim2Real                 & $\mathbf{35.50 \pm 1.87}$  & $\mathbf{18.00 \pm 1.87}$  & $\mathbf{3.50 \pm 3.39}$ & $\mathbf{19.00 \pm 2.47}$ \\
        \bottomrule
    \end{tabular}%
    }
\end{table}

\begin{table}[t]
    \centering
    \caption{List of system identification results for each band type. The rightmost column reports the total loss $\mathcal{L}$ in Eq.~(\ref{eq_loss}), computed using the calibrated simulator. }
    \label{si_loss}

    % Resize the table to fit column width
    \resizebox{\columnwidth}{!}{%
    \begin{tabular}{clcccccc}
        \toprule
        \textbf{Material} & \textbf{Method} & \textbf{\shortstack{Young's\\Modulus}} & \textbf{\shortstack{Poisson's\\Ratio}} & \textbf{\shortstack{Elasticity\\Damping}} & \textbf{\shortstack{Dynamic\\Friction}} & \textbf{\shortstack{Damping\\Scale}} & \textbf{\shortstack{Loss $\downarrow$ \\ ($\mathcal{L}$)}} \\
        \midrule

        % ---------------- BAND 1 ----------------
        \multirow{3}{*}{\textbf{\shortstack{Band 1\\(Soft)}}} 
        & Real2Sim2Real~\cite{real2sim2real}          & 8.52E+6 & 0.45 & 5.12E-3 & 0.94 & 1.00 & 4.1798 \\
        & Package SI~\cite{paramLearning4packages} & 7.06E+5 & 0.45 & 1.30E-2 & 0.25 & 1.00 & 2.4718 \\
        & Sling2Sim2Real                 & 9.29E+5 & 0.46 & 1.11E-2 & 0.80 & 0.79 & \textbf{1.9785} \\
        \midrule
        
        % ---------------- BAND 2 ----------------
        \multirow{3}{*}{\textbf{\shortstack{Band 2\\(Medium)}}} 
        & Real2Sim2Real~\cite{real2sim2real}          & 1.63E+5 & 0.45 & 5.13E-3 & 0.61 & 1.00 & 4.2244 \\
        & Package SI~\cite{paramLearning4packages} & 2.23E+6 & 0.45 & 9.38E-3 & 0.25 & 1.00 & 2.3267\\
        & Sling2Sim2Real                 & 5.01E+6 & 0.38 & 1.01E-2 & 0.28 & 0.77 & \textbf{2.1749} \\
        \midrule

        % ---------------- BAND 3 ----------------
        \multirow{3}{*}{\textbf{\shortstack{Band 3\\(Stiff)}}} 
        & Real2Sim2Real~\cite{real2sim2real}          & 1.27E+5 & 0.45 & 5.13E-3 & 0.98 & 1.00 & 3.7900 \\
        & Package SI~\cite{paramLearning4packages} & 9.56E+5 & 0.45 & 2.43E-2 & 0.25 & 1.00 & 3.4290 \\
        & Sling2Sim2Real                & 3.33E+6 & 0.40 & 7.89E-3 & 0.40 & 0.71 & \textbf{0.9416} \\
        \bottomrule
    \end{tabular}%
    }
\end{table}

\begin{table}[t]
    \centering
    \caption{Ablation study on the effect of the optimization methodology in Sling2Sim2Real.}
    \label{tab:ablation_optimization}
    
    \resizebox{\columnwidth}{!}{%
    \begin{tabular}{llcccc} 
        \toprule
        \multirow{2}{*}{\textbf{Material}} & \multirow{2}{*}{\textbf{Method}} & \multicolumn{4}{c}{\textbf{Landing Distance Error (\unit{cm}) $\downarrow$}}\\
        \cmidrule(lr){3-6}
        & & $g=172.5$ & $g=192.5$ & $g=212.5$ & All Targets \\
        \midrule
        
        \multirow{4}{*}{\shortstack{\textbf{Band 1}\\\textbf{(Soft)}}}
        & $DE_{\mathcal{L}}$   & $19.00 \pm 1.22$ & $11.50 \pm 1.22$ & $\mathbf{3.00 \pm 1.87}$ & $11.17 \pm 0.85$ \\
        & $BO_{\mathcal{L}}$   & $16.50 \pm 1.22$ & $3.00 \pm 1.87$ & $8.00 \pm 6.00$ & $9.17 \pm 2.12$ \\
        & $Adam_{\mathcal{L}}$ & $11.00 \pm 2.00$ & $20.00 \pm 3.16$ & $34.50 \pm 5.79$ & $21.83 \pm 2.31$ \\
        \cmidrule(lr){2-6}
        & Sling2Sim2Real & $\mathbf{11.00 \pm 2.00}$  & $\mathbf{1.00 \pm 1.22}$  & $9.50 \pm 4.30$ & $\mathbf{7.17 \pm 2.68}$ \\
        \bottomrule
    \end{tabular}%
    }
\end{table}

\begin{table}[t]
    \centering
    \caption{Ablation study on the effect of covariance-informed multi-start initialization and the mechanical work loss formulation. The superscripts $-c$, $-m$, and $-\mathcal{L}_\text{work}$ denote the removal of covariance-informed initialization, the multi-start scheme, and the mechanical work loss from Sling2Sim2Real, respectively. }
    \label{tab:ablation_combined}
    
    \resizebox{\columnwidth}{!}{%
    \begin{tabular}{llcccc} 
        \toprule
        \multirow{2}{*}{\textbf{Material}} & \multirow{2}{*}{\textbf{Method}} & \multicolumn{4}{c}{\textbf{Landing Distance Error (\unit{cm}) $\downarrow$}}\\
        \cmidrule(lr){3-6}
        & & $g=172.5$ & $g=192.5$ & $g=212.5$ & All Targets\\
        \midrule
        
        % Example using Band 3 as the material, matching your original ablation caption
        \multirow{4}{*}{\shortstack{\textbf{Band 1}\\\textbf{(Soft)}}} 
        
        % --- Loss Formulation Section ---
        %& \multicolumn{5}{l}{\textbf{Loss Formulation}} \\
        %\cmidrule(lr){2-6}
        %&Sling2Sim2Real$^{-\mathcal{L}_{\text{work}}}$   & $33.00 \pm 1.00$ & $30.00 \pm 1.58$ & $20.50 \pm 2.45$ & $27.83 \pm 5.62$ \\
        %\cmidrule(lr){2-6}
        
        % --- Search Strategy Section ---
        %& \multicolumn{5}{l}{\textbf{Search Strategy}} \\
        %\cmidrule(lr){2-6}
        & Sling2Sim2Real$^{-c,-m}$                & $15.00 \pm 1.58$ & $6.00 \pm 1.22$ & $\mathbf{0.50 \pm 2.45}$ & $7.83 \pm 5.39$ \\
        & Sling2Sim2Real$^{-m}$           & ${29.50 \pm 2.92}$ & $15.00 \pm 3.87$ & $16.50 \pm 2.00$ & $20.33\pm 7.18$ \\
        & Sling2Sim2Real$^{-c}$           & ${22.50 \pm 3.87}$ & $\mathbf{1.00 \pm 1.22}$ & $6.50 \pm 5.39$ & $10.00 \pm 9.91$ \\   
        &Sling2Sim2Real$^{-\mathcal{L}_{\text{work}}}$   & $33.00 \pm 1.00$ & $30.00 \pm 1.58$ & $20.50 \pm 2.45$ & $27.83 \pm 5.62$ \\
        \cmidrule(lr){2-6}
        
        % --- The Single 'Ours' Row ---
        & Sling2Sim2Real                     &  $\mathbf{11.00 \pm 2.00}$  & $\mathbf{1.00 \pm 1.22}$  & $9.50 \pm 4.30$ & $\mathbf{7.17 \pm 2.68}$ \\
        \bottomrule
    \end{tabular}%
    }
\end{table}

% Ablation study
\noindent\textbf{Ablation studies.}
We conduct an ablation study to analyze the impact of our optimization strategy. Table \ref{tab:ablation_optimization} shows that Sling2Sim2Real outperforms baselines that rely on individual optimization techniques. Our two-step optimization improves performance by refining the results of global exploration through multi-start initializations. For the case of $g=212.5\unit{\cm}$, Sling2Sim2Real exhibits a slightly higher error, indicating that the CMA-ES refinement initialized from DE may degrade performance. This result suggests that, although the overall optimization scheme generally provides superior performance, the designed loss and local refinement do not always guarantee accurate Sim2Real transfer. This limitation likely arises from simulation inaccuracies about elastic objects and the highly non-linear parameter space across target configurations. As future work, we plan to investigate target-specific calibration and large-scale calibration across varying targets.
%are required for optimal performance. We remain these for our future work.

We also investigate the effectiveness of our covariance-informed multi-start optimization. As shown in the rightmost column of Table~\ref{tab:ablation_combined}, Sling2Sim2Real outperforms all ablation variants: the single-start method Sling2Sim2Real$^{-m}$, the single-start method without covariance-informed initialization Sling2Sim2Real$^{-c,-m}$, and the multi-start method without covariance-informed initialization Sling2Sim2Real$^{-c}$. This result indicates that combining covariance-informed initialization with a multi-start optimization strategy is effective for Real2Sim SI. Our method also significantly outperforms Sling2Sim2Real$^{-\mathcal{L}_\text{work}}$, demonstrating that the mechanical work loss $\mathcal{L}_\text{work}$---particularly the force-based term---substantially improves SI performance. This result highlights the importance of incorporating real-to-sim force discrepancies, which are not considered in existing Real2Sim SI methods. 
%\da{covariance!}
%single-start ablation methods, Sling2Sim2Real$^{-m}$, that use a DE-induced parameter set with or without empirically measured covariance, as an initial optimization point for CMA-ES.

\noindent\textbf{Qualitative studies.} We finally demonstrate the slingshot policy learning and transfer based on simulated SI. Fig.~\ref{fig:rl_comparison_visual} shows the projectile trajectories after launch produced by three methods: Sling2Sim2Real, Package SI \cite{paramLearning4packages}, and Real2Sim2Real \cite{real2sim2real}. The overlaid projectile trajectories indicate Sling2Sim2Real accurately launches the projectile using unforeseen sling toward the distant target without real-world trial-and-error learning.

\section{Conclusion}
We proposed Sling2Sim2Real, a one-shot Real2Sim2Real policy-learning framework for elastic object manipulation. The framework identifies highly nonlinear and visually indistinguishable physical parameters of elastic objects from a single non-destructive real-world interaction. To address the challenging parameter space, we employ a covariance-informed multi-start optimization strategy that captures correlations between five elasticity parameters, including \textit{damping}, through a two-stage hierarchical optimization process. Using the calibrated simulator, we train a reinforcement learning agent to learn a robust slingshot manipulation policy and transfer it directly to the real world without additional adaptation. Experimental results demonstrate that Sling2Sim2Real outperforms state-of-the-art baselines in terms of landing distance error across diverse elastic bands and target distances. Furthermore, we successfully demonstrate zero-shot sim-to-real policy transfer on real-world slingshot manipulation tasks.

\bibliographystyle{ieeetr}
\nocite{*}   %Cite Everything for now (remove later)
\bibliography{references}

\end{document}